\documentclass[10pt,twocolumn,letterpaper]{article}

\usepackage{iccv}
\usepackage{times}
\usepackage{epsfig}
\usepackage{graphicx}
\usepackage{amsmath}
\usepackage{amssymb}

\usepackage{subfiles}
\usepackage{multirow}
\usepackage{graphics}
\usepackage[export]{adjustbox}
\usepackage[ruled,linesnumbered]{algorithm2e}
\usepackage{booktabs}
\usepackage{makecell}
\usepackage{subfiles}
\usepackage[table]{xcolor}
\usepackage{diagbox}
\usepackage{subfigure}
\usepackage{authblk}
\usepackage[pagebackref=true,breaklinks=true,colorlinks,bookmarks=false]{hyperref}

\iccvfinalcopy 

\def\iccvPaperID{11852} 

\begin{document}

\title{Rethinking Data Distillation: Do Not Overlook Calibration}

\renewcommand\Authands{, }
\makeatletter
\renewcommand\AB@affilsepx{, \protect\Affilfont}
\makeatother

\author[1]{Dongyao Zhu}
\author[2]{Bowen Lei}
\author[3]{Jie Zhang}
\author[4]{Yanbo Fang}
\author[5]{Yiqun Xie}
\author[6]{Ruqi Zhang}
\author[*]{Dongkuan Xu}

\affil[1]{Unaffiliated}
\affil[2]{Texas A\&M University}
\affil[3]{Zhejiang University}
\affil[4]{Certik}
\affil[5]{University of Maryland, College Park}
\affil[6]{Purdue University}
\affil[*]{North Carolina State University}


\maketitle

\begin{abstract}
Neural networks trained on distilled data often produce over-confident output and require correction by calibration methods. Existing calibration methods such as temperature scaling and mixup work well for networks trained on original large-scale data. However, we find that these methods fail to calibrate networks trained on data distilled from large source datasets. In this paper, we show that distilled data lead to networks that are not calibratable due to (i) a more concentrated distribution of the maximum logits and (ii) the loss of information that is semantically meaningful but unrelated to classification tasks. To address this problem, we propose Masked Temperature Scaling (MTS) and Masked Distillation Training (MDT) which mitigate the limitations of distilled data and achieve better calibration results while maintaining the efficiency of dataset distillation. Our code is available upon request.
\end{abstract}

\section{Introduction}

Dataset distillation (DD) has recently gained growing attention because of its ability to reduce the need for large amounts of data during deep neural network (DNN) training, thereby reducing training time and storage burden~\cite{wang2018dataset}. Despite the efficiency of training, studies have pointed out that DD still has multiple limitations. On the one hand, the distillation process is found to be time-consuming, computationally expensive, and storage intensive~\cite{wang2018dataset,zhao2020dataset,zhao2021dataset,deng2022remember,nguyen2020dataset,nguyen2021dataset,kim2022dataset,zhang2023addressing}. On the other hand, DNNs trained on DD data are said to be poorly generalizable to different models or downstream tasks~\cite{wang2018dataset, zhao2020dataset, zhao2021dataset}. Efforts have been conducted to address these issues~\cite{cazenavette2022dataset,zhang2022accelerating,loo2022efficient}.
However, the calibration of DD has been overlooked, which is important for deploying DD safely in real-world applications.

An increasing number of studies are investigating calibration as an important property of DNNs, which means that a DNN should know when it is likely to be wrong~\cite{guo2017calibration, muller2019does, abdar2021review}. In other words, the confidence~(probability related to the predicted category label) of a model should reflect its ground truth correctness likelihood~(accuracy). Previous work has found that DNNs are often too confident to realize when they are making mistakes~\cite{guo2017calibration, ovadia2019can}, which leads to safety issues, especially in safety-critical tasks, e.g., automated healthcare and self-driving cars~\cite{de2022toward, rasheed2022explainable}.

We for the \emph{first} time identify and study the calibration problem of DNNs trained on distilled data (DDNNs).

\begin{figure}[!t]
    \centering
    \scalebox{0.465}{\includegraphics[]{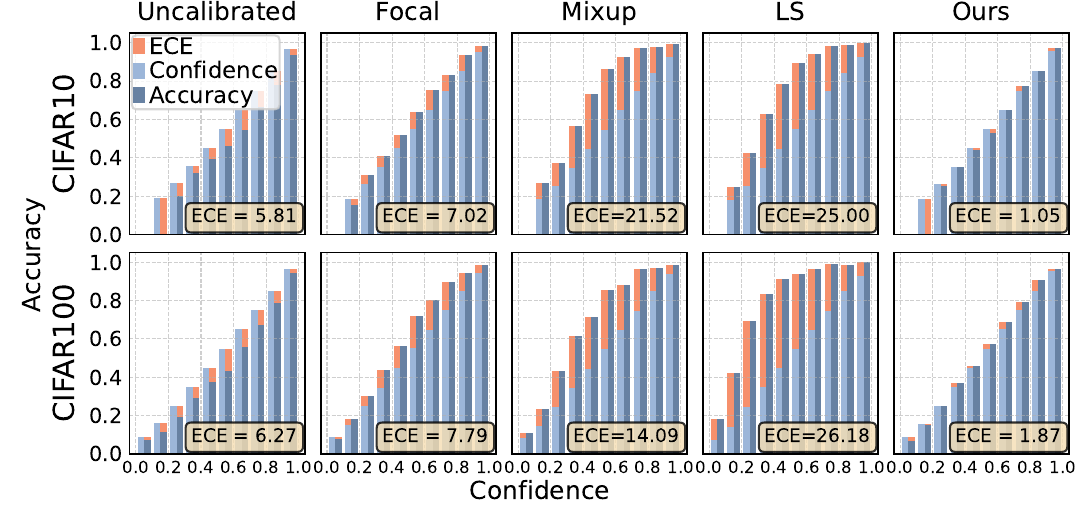}}
    \caption{ECE (red area, smaller is better) of different calibrations on an over-confident ConvNet trained on MTT~\cite{cazenavette2022dataset} distilled CIFAR10 and CIFAR100. Our proposed techniques achieve the best calibration results compared to the over-calibration of other methods. Focal: Focal loss. LS: Label Smoothing.}
    \label{fig:performance_comparison}
    \vspace{-0.5cm}
\end{figure}

\textbf{\textit{Problem 1.}} \textit{We find that DDNNs still suffer from over-confidence problem.} 

We evaluate the calibration quality of DDNNs by Expected Calibration Error (ECE)~\cite{guo2017calibration}, which is a common metric to quantitatively measure the difference between confidence and accuracy. Specifically, to calculate the ECE, we categorize the output probability and accuracy into different levels and calculate the average absolute difference. The lower the ECE, the better the calibration.
As shown in Figure~\ref{fig:performance_comparison}, the ECE (red area) of DDNNs is quite visible in the figures of the first column, which means that the probability of DDNNs' output is usually higher than the actual accuracy of its prediction.
Thus, it is desirable to calibrate DDNNs for reliable prediction and decision-making.

\textbf{\textit{Problem 2.}} \textit{We find that DDNNs are not calibratable when using existing calibration methods.} 

There are calibration methods designed to align the confidence and accuracy of DNNs trained on full datasets (FDNNs). They either modify loss term during network training~\cite{lin2017focal}, use soft labels~\cite{zhang2017mixup, thulasidasan2019mixup}, or scale down the logits after training~\cite{guo2017calibration}.
However, when training on distilled data, we find that most of the existing methods tend to over-calibrate DDNNs.
As shown in Figure~\ref{fig:performance_comparison}, a DDNN trained on distilled CIFAR10 (the first column) has an initial ECE of 6.17\% (red area). 
After calibrating with focal loss (the second column), mixup (the third column), or label smoothing (the fourth column), the DDNN becomes under-confident with increased ECE of 7.79\%, 14.09\%, and 26.18\% respectively, as shown by the inverted and enlarged red bars. This over-calibration problem also occurs for various distillation methods on common datasets (Table~\ref{tab:mts_ece_table}).

In order to address the issues mentioned above, we raise the following questions:

\textbf{\textit{Question 1.}} \textit{Why are DDNNs not calibratable when using existing calibration methods?}

We first dive deep into the differences between the source full data and the distilled data. We find that the distilled data tend to retain information relevant to the classification task while discarding other distributional information in the full data, which may result in limiting DDNNs to pursuing higher accuracy in the classification task while losing more abilities in latent representation learning of FDNNs~\cite{van2017neural,oord2018representation}. By decomposing distilled and full data into smaller components and studying their corresponding significance to model training accuracy, we show that distilled data contains very condensed information, implying a loss of information and leading to harder during-training calibration.
Then, we also investigate the differences between DDNNs and FDNNs. We observe that DDNNs have a more concentrated distribution of logit values, leading to less room for after-training calibration methods such as temperature scaling.

\textbf{\textit{Question 2.}} \textit{How to calibrate DDNNs efficiently?}

To enable DDNNs to be calibratable, we propose (i) Mask Temperature Scaling and (ii) Masked Distillation Training that can be applied both during and after the training of DDNNs. We design a binary masking method for synthetic input when training for distillation objection, which effectively forces the distillation model to extract richer information from the source dataset into distilled datasets, leading to better encoding abilities and thus better calibration of DDNNs. We also show that our proposed masked temperature scaling better improves after-training calibration results on DDNNs by introducing more dynamics to network outputs. Our proposed techniques thus allow for more powerful and more calibratable DDNNs. We summarize contributions as follows:

\begin{itemize}
    \item We for the \emph{first} time study the calibration of DDNNs and find that DDNNs are not calibratable.
    \item We find that DD discards semantically meaningful information and that DDNNs produce a concentrated logit distribution, which explains the difficulty of calibrating DDNNs.
    \item We propose two masking techniques that can improve the calibration of DDNNs better than existing calibration methods, i.e., masked distillation training and masked temperature scaling. In addition, our proposed techniques can be readily deployed in existing dataset distillation methods with minimal extra cost.
    \item We perform extensive experiments on multiple benchmark datasets, model architectures, and data distillation methods. Our techniques reduce ECE values by up to \textbf{91.05\%} with comparable accuracy.
\end{itemize}

\section{Related Work}

\noindent \textbf{Dataset Distillation.} \label{sec: DD}
First introduced by~\cite{wang2018dataset}, dataset distillation is the task of synthesizing a smaller dataset from a large-scale dataset such as CIFAR100~\cite{krizhevsky2009learning}, so that the network trained on the distilled data has a performance comparable to that of the network trained on the source large-scale data. Recent work has significantly improved the performance of networks trained on distilled data and reduced the computational and time overhead of the distillation process while compressing the dataset size to one image per class~\cite{cazenavette2022dataset,deng2022remember,loo2022efficient,nguyen2020dataset,nguyen2021dataset,zhao2020dataset,zhao2021dataset,Wang_2022_CVPR,zhang2022ideal}.
Dataset distillation problem is treated as a gradient-based hyperparameter optimization~\cite{wang2018dataset}. DC performs distillation by matching the gradients generated from distilled data and full data~\cite{zhao2020dataset}. DSA further improves the results by differentiable Siamese augmentations~\cite{zhao2021dataset}. Other SOTA methods include matching trajectories of each parameter between the training on distilled data and full data~\cite{cazenavette2022dataset}, optimizing soft labels~\cite{sucholutsky2021soft}, minimizing reconstruction errors~\cite{yu2023dataset}, and using neural networks to regress features from synthetic samples to real ones~\cite{zhou2022dataset}. The current focus of DD is on computational expense and training performance, and to the best of our knowledge, the difficulties in calibrating over-confident DDNNs remain untouched.
\\

\begin{figure}[!t]
\vspace{-0.4cm}
    \centering
    \subfigure[Maximum logits produced by DDNNs and FDNNs.]{\label{fig:logit_max}
    {\includegraphics[width=0.22\textwidth,height=0.22\textwidth]{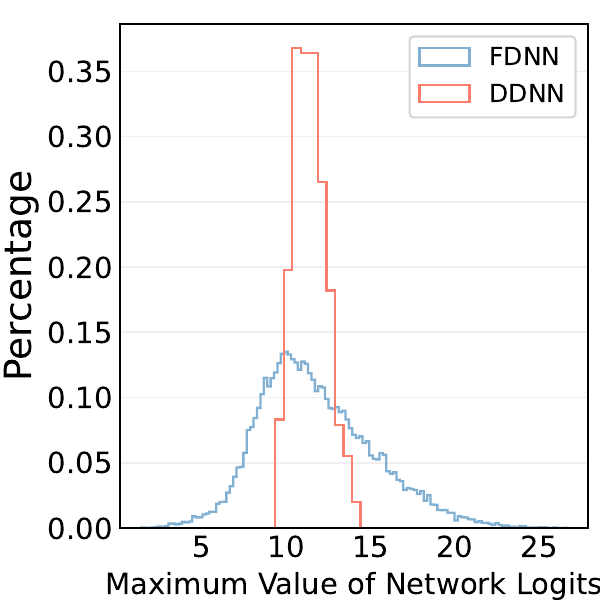}}} \vspace{0cm}
    \subfigure[Prob. of DDNN (top) vs. Ours (bottom) on ID / OOD samples.]{\label{fig:rebuttal_ipc_ood}
    {\includegraphics[width=0.22\textwidth,height=0.215\textwidth]{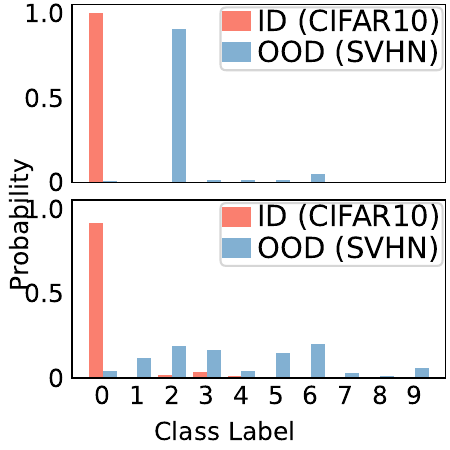}}} 

    \caption{Left: The more calibratable FDNN outputs more evenly distributed logits, while the less calibratable DDNN outputs a more concentrated logit distribution. Top-Right: The less calibratable DDNN struggles to distinguish between an in-distribution (ID) and an out-of-distribution (OOD) sample using its max logits.}
    \vspace{-0.4cm}
\end{figure}

\noindent \textbf{Neural Network Calibration.}
\label{sec:calibration}
 The importance of neural network calibration has been emphasized and received increasing attention~\cite{guo2017calibration}, with the aim of matching the output probability of a neural network (also known as the network output confidence) with the actual accuracy. \cite{guo2017calibration} also introduces the concept of Expected Calibration Error (ECE), which has now become a standard metric for quantitatively measuring calibration quality. A higher ECE implies a poorer calibration of the neural network, while a 0 implies a perfect calibration. Recent calibration methods that have been proposed for networks trained on large-scale datasets include Label Smoothing (LS)~\cite{yuan2020revisiting}, which smooths a one-hot class label with uniform noise during training, forcing the model to learn loose predictions.
Mixup is similar to label smoothing, where different data-label pairs are mixed to form new data points~\cite{thulasidasan2019mixup, zhang2017mixup}. Focal loss (FL), originally designed to address the class imbalance, modifies the traditional cross-entropy loss in classification problems by adding a moderation term, thus allowing the model to focus more on difficult examples that are easily misclassified but difficult to learn~\cite{lin2017focal, mukhoti2020calibrating}. Temperature scaling (TS) is an after-training calibration method applied to fully trained and fixed-weight networks~\cite{guo2017calibration}.
As an extension of Platt scaling~\cite{platt1999probabilistic}, the temperature scaling method scales the output, denoted by $z$, of the last layer of the network with a scaler T before converting it into a probability:
\begin{equation}
    \hat{q_{i}} = \max_{k} \sigma_{softmax}\left (z_{i} / T \right ) ^{(k)}
    z_{i} \in \mathbb{R}^{D}.
    \label{eq:temp_scaling}
\end{equation}

Other work has discussed the necessity~\cite{wu2023towards} and hardness of network calibration~\cite{cheng2022calibrating, ghoshal2022calibrated, zhang2023accelerating}, as well as the degradation of calibration with distribution shift or model size ~\cite{krishnan2020improving, lei2023calibrating}.

\begin{figure}[!t]
    \vspace{-0.4cm}
    \centering
    \subfigure[DDNNs lose more accuracy than FDNNs across DD settings.]{\label{fig:rebuttal_ipc_svd}{\includegraphics[width=0.22\textwidth,height=0.22\textwidth,keepaspectratio,valign=t]{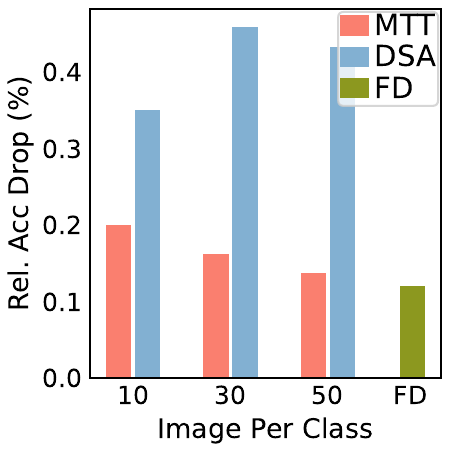}}}
    \subfigure[Trend of accuracy loss vs. dropping singular values (MTT).]{\label{}{\includegraphics[width=0.22\textwidth,height=0.22\textwidth,valign=t]{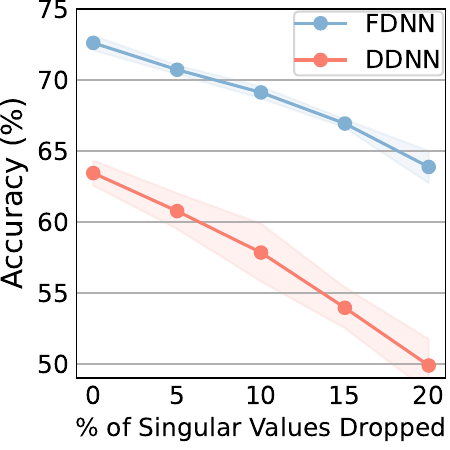}}}
    \caption{Effects on model accuracy of discarding major singular values during SVD reconstruction of distilled and full CIFAR10. DDNNs suffer from more accuracy drop as we discard more singular values during reconstruction, indicating that distilled data contains more condensed information that can be easily grouped by a simple SVD decomposition. Right: IPC = 10.}
    \label{fig:analysis_svd}
    \vspace{-0.4cm}
\end{figure}

\section{Limitation Analysis of DDNNs' Calibration}

We focus on the difficulties of calibrating over-confident DDNNs. As shown in the first column of Figure~\ref{fig:performance_comparison} and the raw ECE reported in Table~\ref{tab:mts_ece_table}, DDNNs show the common over-confidence problem of neural networks, giving higher probabilities than actual accuracy; however, when applied with existing calibration methods, DDNNs are often over-calibrated and become under-confident. In this section, we analyze the reasons that may account for the DDNNs that are not calibratable from 2 aspects: (i) the after-training prediction behaviors and (ii) the during-training network capacity in terms of feature encoding ability. We also discuss the decomposed significance of full data and distilled data on the training accuracy of the network.

\subsection{DDNNs are Less Calibratable}
\label{sec:analysis_network}

We find that the logit distribution of the DDNNs' output is more concentrated, making it difficult to calibrate. In general, a neural network can be considered as a mapping function from the source data domain to the target label distribution, and in the classification task, we use the softmax function to convert logits into label probabilities. The higher the maximum logit value compared to other values, the higher the argmax probability will be and thus the more likely such prediction is over-confident. Therefore we study the distribution of maximum logit values for fully trained DDNNs and FDNNs. As shown in Figure~\ref{fig:logit_max}, the more calibratable FDNNs (blue) output a more dispersed logit distribution, while the less calibratable DDNNs (red) output a concentrated logit distribution with a larger mean.

This mismatched behavior causes problems for after-training calibration methods such as TS and Mixup that operate on scaling output logits, because DDNNs with tight distributions of max logits struggle to distinguish between hard (e.g., out-of-distribution, OOD) and easy (e.g., in-distribution, ID) samples (top of Figure~\ref{fig:rebuttal_ipc_ood}) using the corresponding max logits ~\cite{wei2022mitigating}. Similar theoretical assumptions appear in recent work~\cite{wang2021rethinking}, where they show that the small range of logits due to regularization during training, and a large mean of logits due to the network trying to fit on hard examples may lose information about the different hardness of data points, causing the networks of after-training calibration methods to fail to calibrate.

Therefore, we infer that DDNNs are less calibratable using distilled data due to their more concentrated output distribution and larger mean values. Thus, in order to make DDNNs more calibratable in after-training calibration without modifying network weights, we aim to utilize data that force DDNNs to produce more diverse and smaller outputs.

\subsection{DD Contains Limited Semantic Information}
\label{sec:analysis_svd}

By reconstructing distilled and full data with SVD, we find that distilled data contains only condensed information about the classification task, resulting in the limited ability of DDNNs in latent representation learning.
Intuitively, distilled images should be more informative, or more representative than source full images, in order to keep the number of images small. But do distilled images discard too much source information that is not so much useful for the classification tasks they are optimized for? We hypothesize that distilled data is "simpler" than source full data, such that dropping the same amount of information from distilled datasets should hurt the training performance worse than it does on full data. We start by breaking down full datasets into smaller components of different significance. Singular value decomposition (SVD)~\cite{klema1980singular} is a powerful algorithm in linear algebra for matrix approximation:
\begin{equation}
    U, \Sigma, V = \operatorname{SVD}(X),
    \label{eq:svd_forward}
\end{equation}
where higher singular values in $\Sigma$ correspond to more significant components of $X$. Source data can then be approximately reconstructed by
\begin{equation}
    X^{\prime} \approx U \cdot \Sigma^{\prime} \cdot V^{T}
    \label{eq:svd_backward}
\end{equation}
SVD has been widely used in DNN research for model reconstruction~\cite{xue2013restructuring, xue2014singular}, knowledge distillation~\cite{lee2018self}, and analyzing data~\cite{henry19928}. For our purposes of analyzing data information diversity and significance of data components, we gradually throw away the highest singular values during SVD reconstruction and check for accuracy drop when trained on the approximately reconstructed data. 

Our assumption is that distilled data contains dense information that can be easily grouped, such that SVD decomposes distilled data into several important components and other very small components, compared to full data components whose importance can be more evenly distributed, such that dropping the same number of important components from distilled data would lose more accuracy than in full data.
We drop from 0\% to 20\% of the highest singular values from full CIFAR10 and CIFAR10 distilled by~\cite{cazenavette2022dataset} with IPC = 10, 30 and 50, then train a ConvNet for 300 epochs on the resulting data. As shown in Figure~\ref{fig:analysis_svd}, DDNNs suffer much more severely from the loss of principle components than FDNNs. 

We thus conclude that the distilled data discards too much other semantically meaningful information from the original full data due to over-optimization of the classification task, resulting in condensed information that can be easily decomposed by SVD.

\begin{figure}[!t]
    \vspace{-0.2cm}
    \centering
    \scalebox{1.1}[1.1]{\includegraphics[width=\columnwidth]{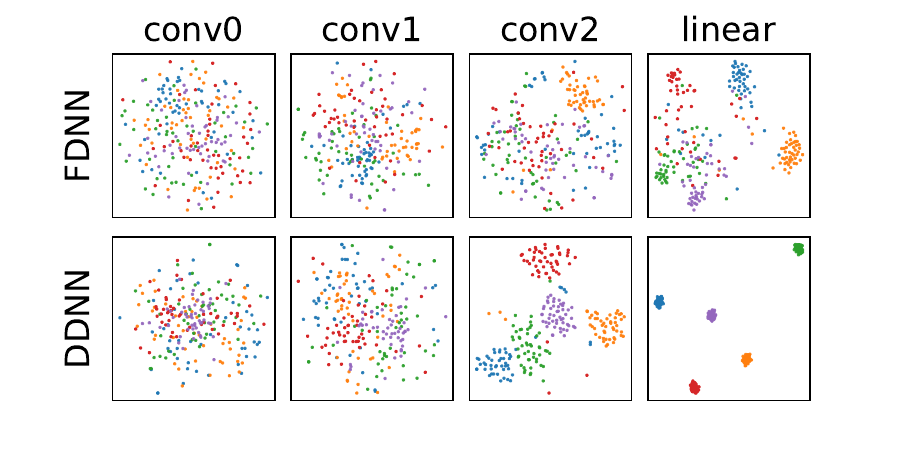}}
    \caption{T-SNE projections of feature vectors from each layer of a 4-block ConvNet trained with Mixup on distilled and full data. FDNNs better encode source information as visualized by the rich features not separated until the last layer. DDNN poorly encodes source information, as shown by the feature projections already separated in layer conv2.}
    \label{fig:analysis_proj}
    \vspace{-0.4cm}
\end{figure}

\begin{table*}[!t]
    \vspace{-0.2cm}
    \addtolength{\tabcolsep}{0pt}
    \renewcommand{\arraystretch}{0.95}
    \centering
    \caption{ECE (\%) of different calibration methods on DDNNs with different DD backbones. Our proposed method yields the best or comparable ECE results in all distillation settings, reducing ECE of DSA by 91.05\%. More importantly, our method does not over-calibrate DDNNs as other calibration methods do (italic), as counted in the last row. Although MX and LS outperform our method in distillation backbones of inferior accuracy (RTP, DC), we surpass them by fine-tuning a more aggressive $r$, as described in Section~\ref{sec:ablations}.}
    \label{tab:mts_ece_table}
    
    \setlength{\aboverulesep}{0pt}
    \setlength{\belowrulesep}{0pt}
    \setlength{\extrarowheight}{.75ex}
    
    \begin{tabular}{cl|ccccc>{\columncolor[gray]{0.9}}cc}
    \toprule
    \multicolumn{2}{c|}{DD Backbone}                            & Raw                  & TS                            & MX                      & LS                        & FL                         & Ours                          \\ \hline 
    \multicolumn{1}{c}{\multirow{4}{*}{MTT}} & CIFAR10         & 4.93 $\pm$ 0.2      & \textit{7.45 $\pm$ 2.1}      & \hspace{-0.08in}\textit{21.06 $\pm$ 0.8}  & \hspace{-0.08in}\textit{25.01 $\pm$ 0.2} & \textit{6.62 $\pm$ 0.2}   & \textbf{1.20 $\pm$ 0.3}      \\  
    \multicolumn{1}{c}{}                     & CIFAR100        & 5.95 $\pm$ 0.4      & \textit{7.76 $\pm$ 0.4}      & \hspace{-0.08in}\textit{14.19 $\pm$ 0.4}  & \hspace{-0.08in}\textit{26.36 $\pm$ 0.4} & \textit{8.30 $\pm$ 0.5}   & \textbf{2.18 $\pm$ 0.2}      \\
    \multicolumn{1}{c}{}                     & Tiny ImageNet           & \hspace{-0.08in}15.78 $\pm$ 0.3     & 2.44 $\pm$ 0.3              & 2.42 $\pm$ 0.3            & \hspace{-0.08in}12.14 $\pm$ 0.3          & 3.61 $\pm$ 0.3            & \textbf{2.26 $\pm$ 0.3}      \\
    \multicolumn{1}{c}{}                     & ImageNette           & 8.68 $\pm$ 1.9      & 4.85 $\pm$ 0.6               & 5.19 $\pm$ 0.6            & \hspace{-0.08in}\textit{23.45 $\pm$ 1.4} & 6.87 $\pm$ 1.3            & \textbf{4.78 $\pm$ 0.5}      \\ \hline 
    \multicolumn{1}{c}{\multirow{2}{*}{RTP}} & CIFAR10         & 2.96 $\pm$ 0.5     & \textit{3.28 $\pm$ 0.7}     & \hspace{-0.08in}\textit{13.35 $\pm$ 1.5}  & \textit{9.58 $\pm$ 0.5} & \textit{8.35 $\pm$ 1.4}   & \textbf{2.22 $\pm$ 0.5}               \\ 
    \multicolumn{1}{c}{}                     & CIFAR100        & \hspace{-0.08in}29.71 $\pm$ 0.6     & \hspace{-0.08in}23.72 $\pm$ 0.6              & \textbf{3.55 $\pm$ 0.6}   & 7.94 $\pm$ 0.2           & \hspace{-0.08in}18.51 $\pm$ 0.5           & \hspace{-0.08in}10.14 $\pm$ 0.4              \\ \hline 
    \multicolumn{1}{c}{DC}                   & CIFAR10         & \hspace{-0.08in}23.60 $\pm$ 0.7     & 5.00 $\pm$ 0.7              & 1.83 $\pm$ 0.3            & \textbf{1.28 $\pm$ 0.1}  & \hspace{-0.08in}13.31 $\pm$ 0.9           & \hspace{-0.08in}10.39 $\pm$ 0.8              \\ \hline 
    \multicolumn{1}{c}{DSA}                  & CIFAR10         & \hspace{-0.08in}19.91 $\pm$ 0.3     & 1.95 $\pm$ 0.4              & 6.44 $\pm$ 0.8            & 2.32 $\pm$ 0.5  & 7.95 $\pm$ 0.7            & \textbf{1.70 $\pm$ 0.4}               \\ \hline   
    \rowcolor[gray]{0.9}\multicolumn{2}{c|}{\textbf{\textit{\# over-calibration}}}          & -                    & 3                            & 3                          & 4                         & 3                          & \textbf{0}                    \\                        
    \bottomrule
    \end{tabular}
    \vspace{-0.2cm}
\end{table*}

\subsection{Limited Semantic Information Weakens Encoding Capacity}
\label{sec:analysis_feature}

We further infer that DDNNs may be less capable of tasks other than classification due to the likely loss of non-classification information. Usually, outputs of intermediate layers of DNNs could be used as feature vectors for other interesting non-classification tasks such as style transfer~\cite{gatys2016image,johnson2016perceptual} due to their unique encodings of source information. To see this, we visualize outputs of layers at different depths in the ConvNet using t-SNE that projects feature vectors down to 2 dimensions. We can see that in Figure~\ref{fig:analysis_proj}, features from FDNNs cluster slowly, and become visually separable only in the last layer, thus retaining most of the original information in its latent vectors; features from DDNNS, however, form visible cluster already in layer conv2, making more compact final clusters that are more valuable for classifications than other tasks such as feature extraction. Moreover, clusters of DDNNs from each class are closer to each other than those from FDNNs. Clearly, outputs of middle layers from DDNNs are already alike label distributions, discarding too much non-classification information. Similar observations on other distillation backbones are reported in~\cite{Wang_2022_CVPR}, in which they also account for long-tailed gradients as possible reasons. 

Therefore, DDNNs do not exhibit good encoding capability due to being trained on distilled data optimized specifically for the classification task and may be susceptible to being over-calibrated by calibration methods. We provide more details in the supplementary material.

\section{Our Proposed Techniques}

We respond to the analyses in Sections~\ref{sec:analysis_network}-\ref{sec:analysis_feature} so that our method can be applied during and after training, providing calibration options at different times and computational budget levels.

\subsection{Masked Temperature Scaling}

As discussed in Section~\ref{sec:analysis_network}, compared to FDNNs, DDNNs produce a more concentrated distribution of logit values with larger values, and these large and condensed logit values lead to networks that are not calibratable.
Since after-training calibration methods such as temperature scaling \cite{guo2017calibration} make use of these large and concentrated logit values from a forward pass of validation data, we seek to overcome this source of difficulty in calibration by perturbing the validation data such that the model could output more various and smaller logit values. Inspired by dropout \cite{srivastava2014dropout}, we apply a simple zero-masking on the validation data of temperature scaling. Our proposed method, which we refer to as Masked Temperature Scaling (MTS), thus modifies Eq~\eqref{eq:temp_scaling} as follows: 
\begin{equation}
    \hat{q_{i}} = \max_{k} \sigma_{softmax}\left (z_{i} * mask / T \right ) ^{(k)},
\end{equation}
\noindent where q, z, mask $\in R^{D}$ and the number of zeros in the mask is controlled by a hyperparameter masking ratio $r$. Note that masking is only applied when updating $T$, such that MTS does not change model accuracy. We use a sampled portion of the training data we have to update the temperature parameter $T$, instead of using separate validation data as in traditional temperature scaling. 

This is particularly necessary in the dataset refinement setting, as we may simply not have any extra data, for example, when each class of images is set to 1 (see Section~\ref{sec:ablations} for more details).

\subsection{Masked Distillation Training}

In response to the analyses in Sections~\ref{sec:analysis_network}-\ref{sec:analysis_feature}, we avoid over-concentration of distillation data on easily identifiable information in the source complete data by perturbing the binary mask during distillation, so that the distillation data also contain more semantically complete information.

\begin{algorithm}[t]
\caption{Masked Distillation Training}\label{alg:paradigm}
\SetNoFillComment
\SetKwInOut{Output}{Notation}
\SetKwInput{KwResult}{Definition}
\SetKwData{Data}{Definition}
\KwIn{Source training data $\mathcal{T}$, number of \\ classes $N_{c}$, deep neural network $\psi_{\theta}$ parameterized with $\theta$, criterion $C$, loss function $l$, total number of training steps $T$, masking ratio $r$}
\KwOut{Distilled dataset $\mathcal{S}$}
\For{$t\gets0$ \KwTo $T$}{
    custom pre-processing

    \For{$c\gets0$ \KwTo $N_{c}$} {
        Sample $T_{c} \sim \mathcal{T}$, $S_{c} \sim \mathcal{S}$\\
        Update synthetic data $\mathcal{S}_{c}$:\\
        $S_c \leftarrow S_c-\lambda \nabla_{S_c} C\left(S_c, Mask\left( T_c, r \right), l, \theta\right)$
    }
    custom post-processing

    Update $\theta$ of network $\psi_{\theta}$ using $T \sim \mathcal{T}$
}
\end{algorithm}

A typical DD training paradigm tries to minimize the differences of certain characteristics, as measured by some criterion $C$, between data batch $B^{\prime}$ from synthetic data and data batch $B$ from source full data. The loss function $l\left( \theta; X\right)$ used in $C$ is usually the cross entropy loss for training $\theta$ on $x$ in classification tasks. We thus put the binary mask on synthetic data before feeding it into $C$. We give the details of our method, Masked Distillation Training (MDT), in Algorithm~\ref{alg:paradigm}. MDT is applicable to various distillation backbones. For instance, in Efficient Dataset Distillation~\cite{zhang2022accelerating}, they set the criterion $C$ as the differences between gradients back-propagated from $l$ given source data $B$ and distilled data $B^{\prime}$:
\begin{equation}
    C\left(B, B^{\prime} ; l, \theta\right) = \left.\| \nabla_\theta \ell(\theta ; B)-\nabla_\theta \ell\left(\theta ; B^{\prime}\right) \|\right. 
    \\
\end{equation}
 When applied with MDT, this now becomes:
 \begin{equation}
    C\left(B, B^{\prime} ; l, \theta\right) = \left.\| \nabla_\theta \ell(\theta ; B)-\nabla_\theta \ell\left(\theta ; Mask\left( B^{\prime}, r \right)\right) \|\right. 
    \\
\end{equation}

\begin{figure}[!t]
    \vspace{-0.2cm}
    \centering
    \scalebox{0.40}[0.45]{\includegraphics[]{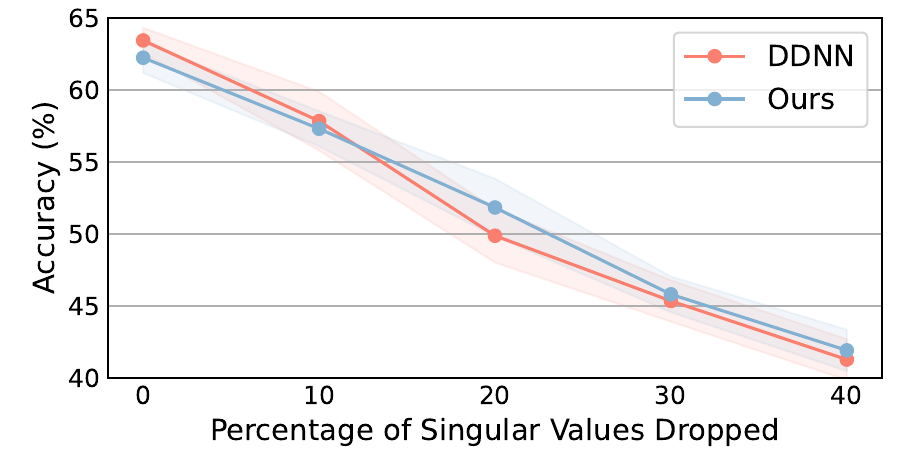}}
    \caption{Model trained on MDT (Ours) distilled data suffers from less accuracy drop than the model (DDNN) trained on MTT distilled data when dropping major singular values during SVD reconstruction of DD, showing that ours alleviates the issue of condensed information of distilled data.}
    \label{fig:result_svd}
    \vspace{-0.2cm}
\end{figure}

\begin{figure}[!t]
    \centering
    \scalebox{1.1}[1.1]{\includegraphics[width=\columnwidth]{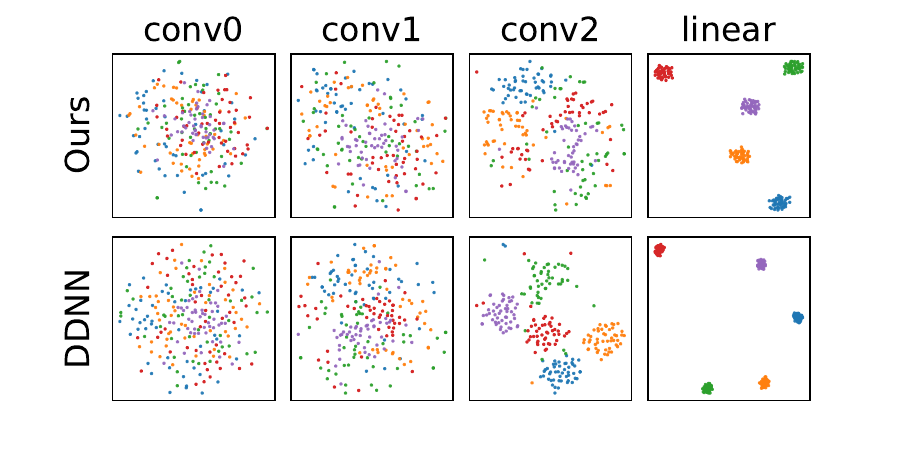}}
    \caption{T-SNE projections of feature vectors from each layer of a 4-block ConvNet trained on original distilled data and Ours. The model trained on Ours better encodes source information than the original DDNN, as visualized by the features that are hardly separated in layer conv2.}
    \label{fig:result_proj}
    \vspace{-0.2cm}
\end{figure}

 In another distillation backbone MTT~\cite{cazenavette2022dataset}, the criterion $C$ measures parameter trajectory differences between DDNNs and FDNNs:
\begin{align}
\begin{split}
    \hat{\theta}_{t+N} \leftarrow \hat{\theta}_{t+N-1}-\lambda \nabla l\left(B^{\prime}, \theta_{t+N-1}\right) \\
    C\left(B^{\prime}, \hat{\theta}; l, \theta\right) = \left \| \hat{\theta}_{t+N} - \theta^{*}_{t+M} \right \| ^{2}_{2} / \left \| \theta^{*}_{t} - \theta^{*}_{t+M} \right \| ^{2}_{2}
\end{split}
\end{align}
and when applied with MDT, this becomes:
\begin{equation}
    \hat{\theta}_{t+N} \leftarrow \hat{\theta}_{t+N-1}-\lambda \nabla l\left(Mask\left( B^{\prime}, r \right), \theta_{t+N-1}\right).
\end{equation}
We find that a masking ratio of 10\% works well for our purposes while losing minimal test accuracy, and we provide more details in Sections \ref{sec:acc_mdt} and \ref{sec:other_mdt}.

\subsection{Connection to Dropouts}

Dropout~\cite{srivastava2014dropout} is a common practice to prevent overfitting of neural networks. Two popular types are unit dropout (U-DP) and weight dropout (W-DP), which randomly discard units (neurons) and individual weights at each training step, respectively. The formulas are shown in Eq~(\ref{eq:dp}).
\begin{align}
    \text{U-DP:} \ Y &= (X \odot M ) W; \ \ 
    \text{W-DP:} \ Y = X (W \odot M) \label{eq:dp},
\end{align}
where $M$ denotes dropout mask and $W$ refers to weights.

Our proposed Masked Distillation Training can be viewed as a new version of dropout on the input, i.e., $X = S_c \odot M$. 
There are practices using dropout on inputs as data augmentation~\cite{bouthillier2015dropout}. In contrast to existing efforts, we apply masking in distillation backbones on synthetic data during their forward passes. Masking some of the synthetic data makes it harder to collect easily reachable information from the source dataset, and thus forces the distillation to focus on other structurally and semantically meaningful information that has not received sufficient attention in previous data distillation.

\begin{figure}[!t]
    \vspace{-0.5cm}
    \centering
    \subfigure[Maximum Logits on CIFAR10]{
    {\includegraphics[width=0.23\textwidth,height=0.22\textwidth]{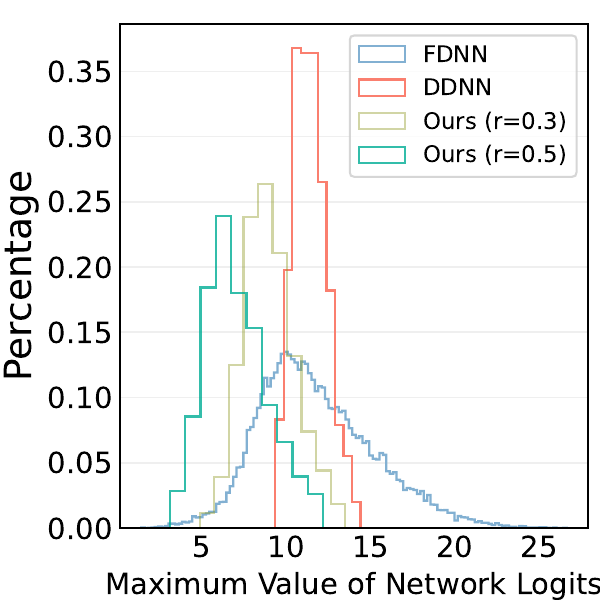}}} \vspace{0cm}
    \subfigure[Maximum Logits on CIFAR100]{
    {\includegraphics[width=0.23\textwidth,height=0.22\textwidth]{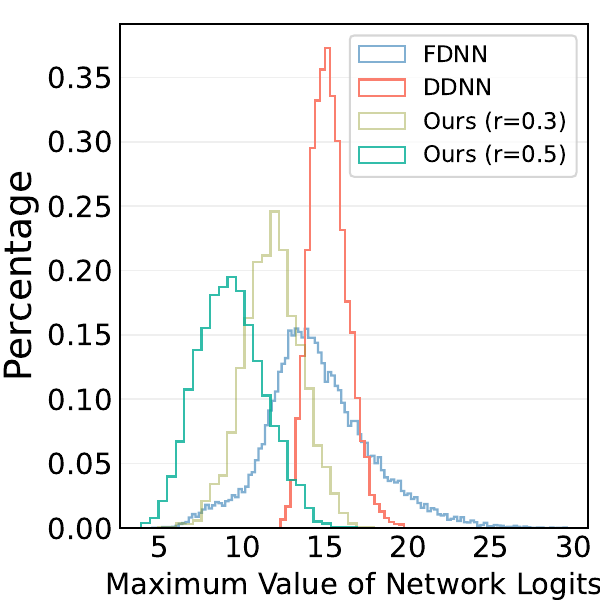}}}
    \caption{Histogram of maximum logits of DDNNs (Ours) on CIFAR10 and CIFAR100. As we increase the ratio of masking in our method, DDNNs produce logits that cover more values, thus becoming more calibratable by after-training calibration methods.}
    \label{fig:result_max}
    \vspace{-0.2cm}
\end{figure}

\section{Experiments}

\subsection{Experiment Setup}

We thoroughly evaluate our proposed MTS and MDT on different dataset distillation setups and compare them with existing calibration methods.
\noindent \textbf{Dataset Distillation Backbones:} We follow the exact settings in MTT~\cite{cazenavette2022dataset}, RTP~\cite{deng2022remember}, DC~\cite{zhao2020dataset} and DSA~\cite{zhao2021dataset}. Our experiments are based on 4 benchmark datasets: CIFAR10 \& CIFAR100~\cite{krizhevsky2009learning}, Tiny ImageNet~\cite{le2015tiny}, and ImageNette (a subset of ImageNet)~\cite{Howard_Imagenette_2019}. We mainly set image-per-class to larger values, e.g. 50 in MTT, and results on different IPCs are provided in supplement materials.
\noindent \textbf{Calibration Methods:} We compare our method with existing calibration methods including Temperature Scaling~(TS)~\cite{guo2017calibration}, mixup~(MX)~\cite{zhang2017mixup}, Label Smoothing~(LS)~\cite{yuan2020revisiting}, and Focal Loss~(FL)~\cite{lin2017focal}. 
\noindent \textbf{Implementations:} For TS, we use an initial temperature of 1.5 and LBFGS \cite{liu1989limited} optimizer with a learning rate of 0.02. For MX, we use a $\beta$ distribution with $\alpha = 1.0$ for the mixup ratio. For LS, we set $\epsilon = 0.1$. For FL, we set $\gamma = 1.$, which calibrates better on DDNNs than the best value 2 reported in the paper. For our proposed MTS, on distillation backbone MTT, we use a fixed masking ratio of 0.3, 0.3, 0.5, and 0.1 for each of the 4 datasets respectively. On backbones RTP, DSA, and DC, due to their inferior performance in accuracy, we use a more aggressive masking ratio of 0.8. 

Since the number of examples of distillation data is usually limited, we draw 10\% of all distillation data as the validation set for the after-training method, as in other existing work. The experiments are repeated five times, and the mean and standard deviation are reported. More experimental setups are available in supplementary materials.

\subsection{Empirical Analysis of MTS}

We show in Figure~\ref{fig:performance_comparison} that our proposed method is able to reduce the ECE (red bars) to almost zero for each confidence bin when using MTT as the distillation backbone on CIFAR10 and CIFAR100.
Although traditional calibration methods such as mixup~\textit{can} perform well, they could also over-calibrate and result in under-confident networks. We visualize the under-confidence in Figure~\ref{fig:performance_comparison}, in which the red bars are enlarged and switched from left to right in each bin. Additional calibration results are reported in Table~\ref{tab:mts_ece_table}, and our proposed MTS gives the best numerical ECE results in almost all settings. 
In real-world settings where no mistakes are allowed, traditional methods are regarded as unsafe due to their potential over-calibration. 

As a contrast, we propose masked temperature scaling, which not only has better performance but does not show any lack of confidence in the results at all and is therefore considered a safer choice.

\begin{table}[t]
    \addtolength{\tabcolsep}{-2pt}
    \renewcommand{\arraystretch}{1}
    \centering
    \caption{ECE (\%) of different calibration methods on distilled datasets trained with our methods. Tiny: Tiny ImageNet. Nette: Nette subset of ImageNet. Our results are in \colorbox[gray]{0.9}{shadow}.}
    \label{tab:mdt_ece_table}
    \scalebox{0.8}{
    \setlength{\aboverulesep}{0pt}
    \setlength{\belowrulesep}{0pt}
    \setlength{\extrarowheight}{.75ex}
    
    \begin{tabular}{l|c>{\columncolor[gray]{0.9}}c>{\columncolor[gray]{0.9}}c>{\columncolor[gray]{0.9}}cc}
    \toprule
    \multicolumn{1}{c|}{Dataset}         & Best of Others               & MDT              & MTS                       & MDT + MTS                \\ \hline 
    \multicolumn{1}{l|}{CIFAR10}         & 3.64 $\pm$ 0.2 (TS)         & 3.66 $\pm$ 0.3  & \textbf{1.20 $\pm$ 0.3}  & 2.50 $\pm$ 0.5          \\      
    \multicolumn{1}{l|}{CIFAR100}        & 5.95 $\pm$ 0.4 (Raw)        & 4.65 $\pm$ 0.3  & 2.18 $\pm$ 0.2           & \textbf{2.00 $\pm$ 0.5} \\   
    \multicolumn{1}{l|}{Tiny}            & 2.42 $\pm$ 0.3 (MX)      & 7.44 $\pm$ 1.4  & \textbf{2.26 $\pm$ 0.3}  & 5.91 $\pm$ 1.4          \\  
    \multicolumn{1}{l|}{Nette}           & 4.85 $\pm$ 0.6 (TS)         & 7.32 $\pm$ 1.7  & \textbf{4.78 $\pm$ 0.5}  & 5.14 $\pm$ 1.2          \\ \bottomrule
    \end{tabular}}
    \vspace{-0.2cm}
\end{table}

\begin{table}
\setlength{\abovecaptionskip}{-0.0cm}
\setlength\tabcolsep{1pt} 
\begin{small}
\centering

    \setlength{\tabcolsep}{3pt}
    \caption{ECE (\%) of MDT with dynamically sampled $r$ and MTS on CIFAR10, MTT with different IPCs.}
    \vspace{0.1cm}
    \centering
    \begin{tabular}{l|cc>{\columncolor[gray]{0.9}}c}
    \toprule
    \multicolumn{1}{c|}{IPC} & MDT\textsuperscript{ds} & MTS & MDT\textsuperscript{ds} + MTS \\ \hline
    \multicolumn{1}{c|}{10} & 1.79 $\pm$ 0.9 & 1.36 $\pm$ 0.4 & \textbf{1.13 $\pm$ 0.2}    \\
    \multicolumn{1}{c|}{50} & 5.10 $\pm$ 0.4 & 1.20 $\pm$ 0.3 & \textbf{1.26 $\pm$ 0.2}    \\
    \bottomrule
    \end{tabular}

    \label{tab:rebuttal_mdt}
\end{small}
\vspace{-0.3cm}
\end{table}

\subsection{Empirical Analysis of MDT}
\label{sec:acc_mdt}

We show that MDT improves the calibration results. As reported in Tables~\ref{tab:mdt_ece_table}-\ref{tab:rebuttal_mdt}, applying MDT alone or combining it with MTS yields comparable or better calibration performance. WE note that naively combining MDT + MTS may increase ECE due to DDNNs
overfitting to the fixed masking ratio in MDT, then being
over-calibrated by MTS. Thus we further improve (Table \ref{tab:rebuttal_mdt}) MDT + MTS by dynamically sampling the $r$ in MDT from 0 to 0.1 (denoted MDT\textsuperscript{ds}) so the resulting DDNNs are more calibratable. We fix $r$ in MTS due to the limited amount of validation data in DD. This indicates that our proposed MDT produces more robust and calibratable DDNNs than the original backbone when sufficient computational resources are available to train the distillation process from the beginning. We use MTT as the distillation backbone.

We find that MDT gives comparable model accuracy albeit altering the distillation process. With a 10\% zero masking during the distillation process, MDT only leads to a loss of as large as 1.26\% in DDNNs' accuracy on CIFAR100 and as low as 0.14\% on Tiny ImageNet. As reported in Table~\ref{tab:mdt_acc_table}, this is even better than traditional during-training calibration methods such as mixup, label smoothing, and focal loss that lead to different model results. 

This suggests that MDT yields better calibration potential at a negligible performance cost, which is desirable in an environment where security is a major concern~\cite{de2022toward,rasheed2022explainable}.

\begin{table*}[!t]
\vspace{-0.1cm}
    \addtolength{\tabcolsep}{0pt}
    \renewcommand{\arraystretch}{0.9}
    \centering
    \caption{Accuracy (\%) of different during-training calibration methods on MTT distilled datasets. While all during-training calibration methods lead to a loss in accuracy, ours loses only as small as 0.14\% at a masking ratio of 10\%. Our results are in \colorbox[gray]{0.9}{shadow}.}
    \label{tab:mdt_acc_table}

    \setlength{\aboverulesep}{0pt}
    \setlength{\belowrulesep}{0pt}
    \setlength{\extrarowheight}{.75ex}
    
    \begin{tabular}{l|c|ccc>{\columncolor[gray]{0.9}}c}
    \toprule
    \multicolumn{1}{c|}{Dataset}         & Raw                 & MX              & LS                       & FL                       & Ours                 \\ \hline
    \multicolumn{1}{l|}{CIFAR10}         & 70.48 $\pm$ 0.2    & 65.50 $\pm$ 0.5   & 67.42 $\pm$ 0.5         & 68.79 $\pm$ 0.5         & \textbf{69.98} $\pm$ 0.4    \\
    \multicolumn{1}{l|}{CIFAR100}        & 47.47 $\pm$ 0.2    & 39.65 $\pm$ 0.3   & 47.02 $\pm$ 0.2         & 46.79 $\pm$ 0.4         & 46.21 $\pm$ 0.4    \\
    \multicolumn{1}{l|}{Tiny ImageNet}            & 27.76 $\pm$ 0.2    & 21.48 $\pm$ 0.4   & 25.76 $\pm$ 0.3        & 27.42 $\pm$ 0.3          & \textbf{27.62} $\pm$ 0.4  \\
    \multicolumn{1}{l|}{ImageNette}           & 63.04 $\pm$ 1.3    & 55.60 $\pm$ 1.0   & 63.40 $\pm$ 0.9        & 61.32 $\pm$ 0.9          & 62.80 $\pm$ 1.2    \\ \bottomrule
    \end{tabular}
    \vspace{-0.1cm}
\end{table*}

\begin{figure}[t]
\vspace{-0.2cm}
\centering
\subfigure[CIFAR10]{
\begin{minipage}[t]{0.48\linewidth}
\centering
\includegraphics[width=1.5in]{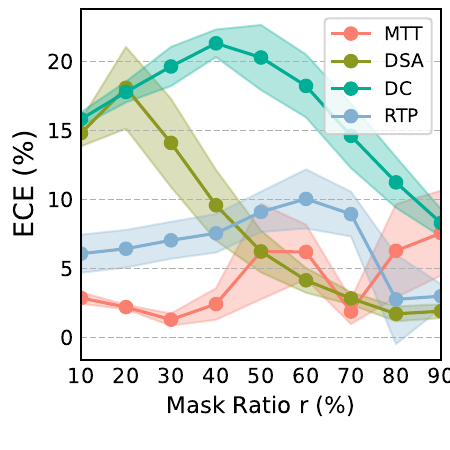}
\label{fig:forgetting_a}
\end{minipage}%
}%
\subfigure[CIFAR100]{
\begin{minipage}[t]{0.48\linewidth}
\centering
\includegraphics[width=1.5in]{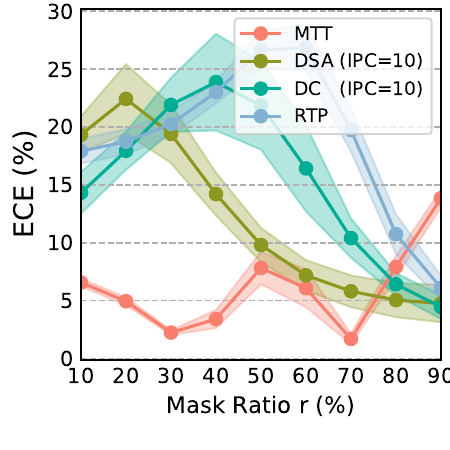}
\end{minipage}%
}%
\centering
\caption{Effects on ECE (\%) of different masking ratios $r$ in our method. For DD methods with better performance reported (MTT, RTP on CIFAR10), our method is robust to $r$ and saves efforts in fine-tuning. For DD methods with inferior performance (DC, DSA, RTP on CIFAR100), a more aggressive masking ratio ($r >$ 0.7) could still calibrate reasonably well.}
\label{fig:ablation_r}
 \vspace{-0.3cm}
\end{figure}

\subsection{Enabling Calibratable DDNNs}
In response to the discussion of the after-training behavior of DDNNs in Section~\ref{sec:analysis_network}, we examined improvements in DDNN calibrability.
On the validation data for Temperature Scaling, we apply zero-masking with ratio $r$ = 10\%, 20\%, and 30\% to see its effects on resulting logit distributions of DDNNs. We show in the bottom-right of Figure \ref{fig:rebuttal_ipc_ood} that our MDT produces lower probabilities on OOD samples, leading to more distinguishable logits and more calibratable DDNNs than before. We also show in Figure \ref{fig:result_max} that DDNNs given these mask-perturbed data will produce similarly diverse logits as if they are processing normal full data, allowing masked temperature scaling to better calibrate DDNNs with similar good performance on FDNNs.

\subsection{Enhancing Semantic Information of DDNNs}
\label{sec:other_mdt}
We investigate whether the semantic information of DD is enhanced according to the discussion in Section~\ref{sec:analysis_svd}. 
As shown in Figure~\ref{fig:result_svd}, when trained with MDT, our DDNNs start with a little lower accuracy than the normal MTT model. However, as we gradually drop more singular values following Eqs~(\ref{eq:svd_forward})-(\ref{eq:svd_backward}), the accuracy of the MDT model drops slower and even stays higher than the accuracy of the MTT models. 
This indicates that MDT distillation effectively retains more semantically meaningful information than normal distillation does, making MDT distilled data more difficult to be decomposed by SVD.

\subsection{Improving Encoding Capacity of DDNNs}

We study the improvement in the feature encoding capability of the DDNNs, responding to the discussion of DDNN behavior during training in Section~\ref{sec:analysis_feature}.
We experiment on the distillation backbone MTT, in which they collect network parameter trajectories from training on synthetic data in each iteration. We apply masked distillation training with masking ratio = 10\% on the synthetic data before the forward pass in each iteration. 

We show in Figure~\ref{fig:result_proj} that the hidden layers in the MDT model form larger clusters than the original MTT model, and that the clusters in each category are more intertwined with each other, retaining more information from the complete dataset and forming better feature vectors as desired.

\section{Ablation Studies}
\label{sec:ablations}

\noindent \textbf{Analysis of Mask Ratio $r$ in MTS.}
We analyze the effects of mask ratio $r$ on the calibration results of MTS. We set $r$ from 0.1 to 0.9, increasing by 0.1. 
As shown in Figure \ref{fig:ablation_r}, on CIFAR10 and CIFAR100, MTS works well for most of the possible $r$ ranging from 0.1 to 0.5, indicating that MTS can be tuned with minimal effort. On variants of the ImageNet dataset, however, we find that 0.3 works best for Tiny ImageNet, and 0.5 for ImageNet Subset. 
This is probably due to the large number of classes in these more complex datasets, as well as the relatively low accuracy of their corresponding distilled datasets.

\begin{figure}[t]
\vspace{-0.1cm}
\centering
\subfigure[CIFAR10]{
\begin{minipage}[t]{0.48\linewidth}
\centering
\includegraphics[width=1.5in]{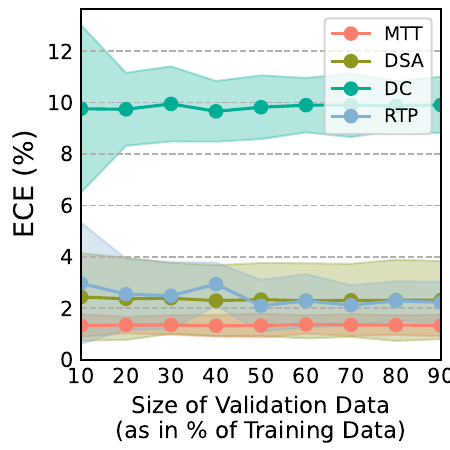}
\label{fig:ablation_len_a}
\end{minipage}%
}%
\subfigure[CIFAR100]{
\begin{minipage}[t]{0.48\linewidth}
\centering
\includegraphics[width=1.5in]{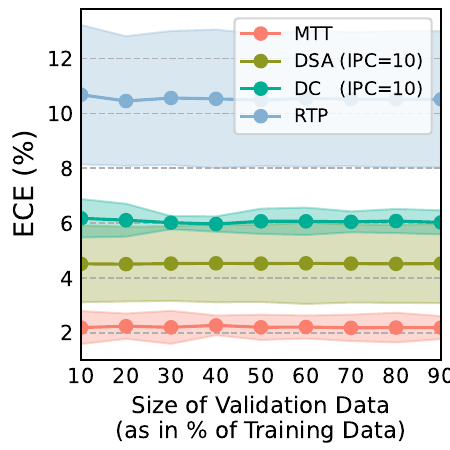}
\end{minipage}%
}%
\centering
\caption{Effects on ECE (\%) of different sizes of the validation data ($N$, as in \% of total training data) in our method. On different distillation backbones, a small $N$ gives identical calibration performance to a larger $N$, indicating that our method is also applicable in scenarios with extremely scarce validation data.}
\label{fig:ablation_len}
\vspace{-0.1cm}
\end{figure}

\noindent \textbf{Analysis of Validation Set Size for MTS.}
We study the impact of how much data is drawn from all the distilled data as validation data for MTS.
We denote $N$ as the proportion we sample, and we set $N$ from 10\% to 50\%, increased by 10\%. We can see in Figure \ref{fig:ablation_len} that the number of samples has little effect on calibration results. We hypothesize that this is due to we only update the temperature parameter $T$ for only one step, thus not being affected by the number of examples in this step. 
This also indicates that MTS can be applied when we have only a small amount of data available, such as distillation with IPC=1 or medical image analysis scenarios~\cite{liu2022deep, chen2022recent, salahuddin2022transparency}.

\noindent \textbf{Calibration in Lower Accuracy Settings.}
Our method outperforms other calibration methods at DD settings with extreme compression ratio, i.e. only 1 synthetic image for each label. This means traditional temperature scaling no longer applies because it requires additional validation data. As reported in Table~\ref{tab:ipc1_table}, while other calibration methods over-calibrate or don't work at all, ours still produces better results, indicating its generality to various DD settings.

\begin{table}[!t]
    \setlength{\tabcolsep}{2.7pt}
    \centering
    \caption{ECE (\%) of different calibration methods with IPC=1. Under this extreme compression rate, our method still outperforms other calibration methods. Our results are in \colorbox[gray]{0.9}{shadow}.}
    \label{tab:ipc1_table}

    \setlength{\aboverulesep}{0pt}
    \setlength{\belowrulesep}{0pt}
    \setlength{\extrarowheight}{.75ex}
    
    \scalebox{0.8}{
    \begin{tabular}{l|cccc>{\columncolor[gray]{0.9}}c}
    \toprule
    \multicolumn{1}{c|}{Dataset}         & Raw                 & MX              & LS                       & FL                       & Ours                 \\ \hline
    \multicolumn{1}{l|}{CIFAR10}     & \hspace{-0.04in}10.15 $\pm$ 1.2    & 8.40 $\pm$ 1.1   & \hspace{-0.04in}12.79 $\pm$ 0.6         & 2.05 $\pm$ 0.9         & \textbf{1.81 $\pm$ 0.7}    \\
    \multicolumn{1}{l|}{CIFAR100}     & 2.46 $\pm$ 0.6    & 4.45 $\pm$ 0.5   & 8.89 $\pm$ 0.6         & 3.24 $\pm$ 0.9         & \textbf{2.19 $\pm$ 0.5}    \\
    \bottomrule
    \end{tabular}
    }
    \vspace{-0.3cm}
\end{table}

\begin{figure}[!t]
    \vspace{-0.4cm}
    \centering

    \subfigure[Mixup]{
    {\includegraphics[width=0.48\textwidth,height=0.22\textwidth]{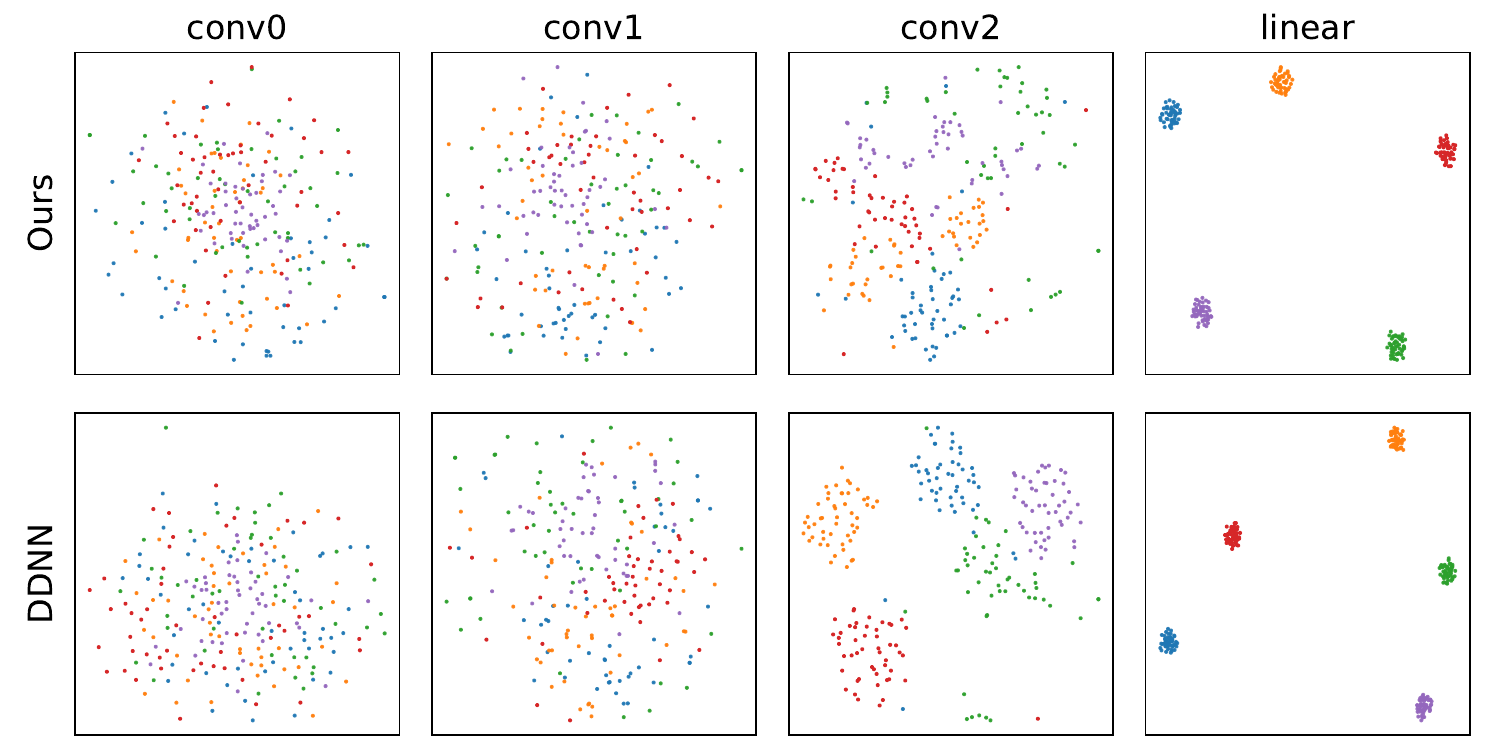}}} \vspace{-0.3cm}

    \subfigure[Label Smoothing]{
    {\includegraphics[width=0.48\textwidth,height=0.22\textwidth]{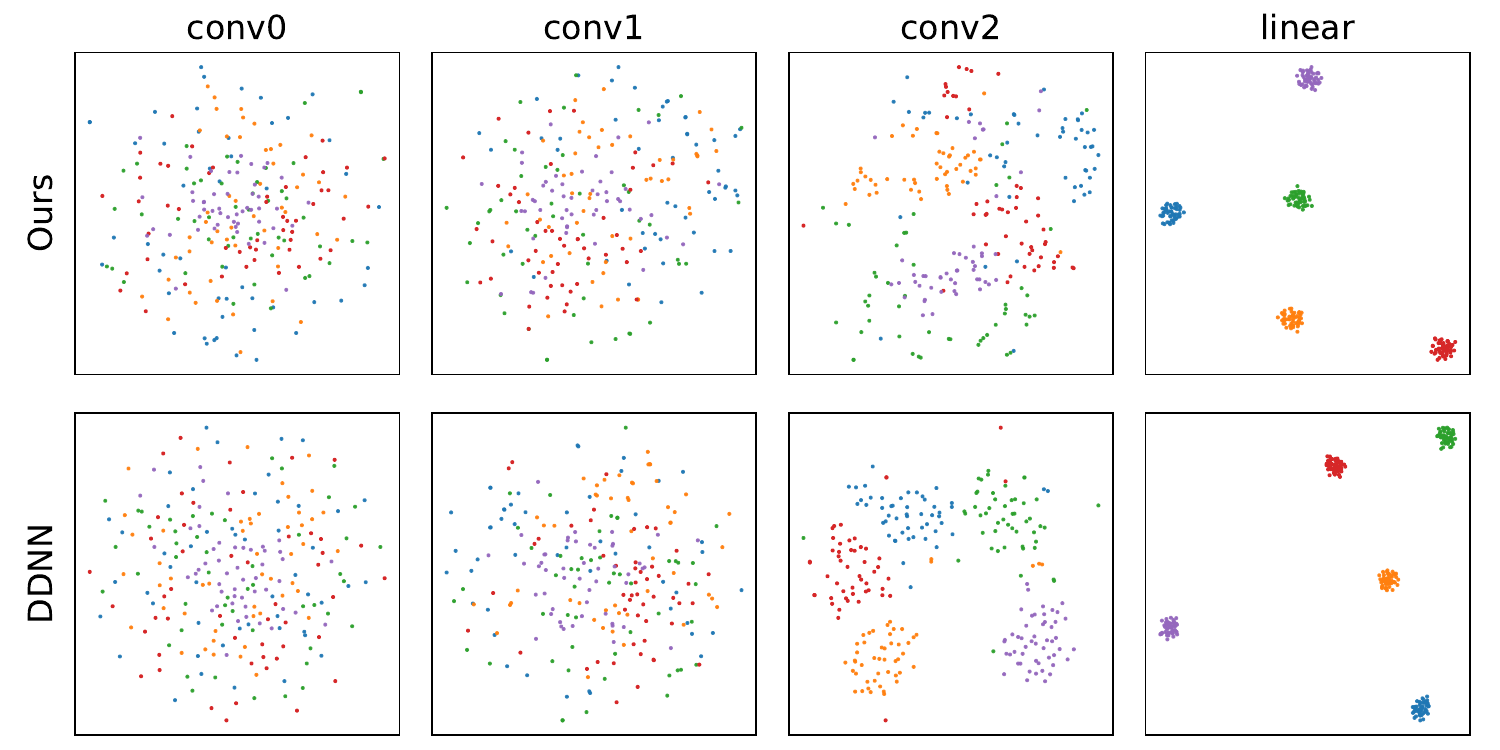}}}
    \vspace{-0.3cm}

    \subfigure[Focal Loss]{
    {\includegraphics[width=0.48\textwidth,height=0.22\textwidth]{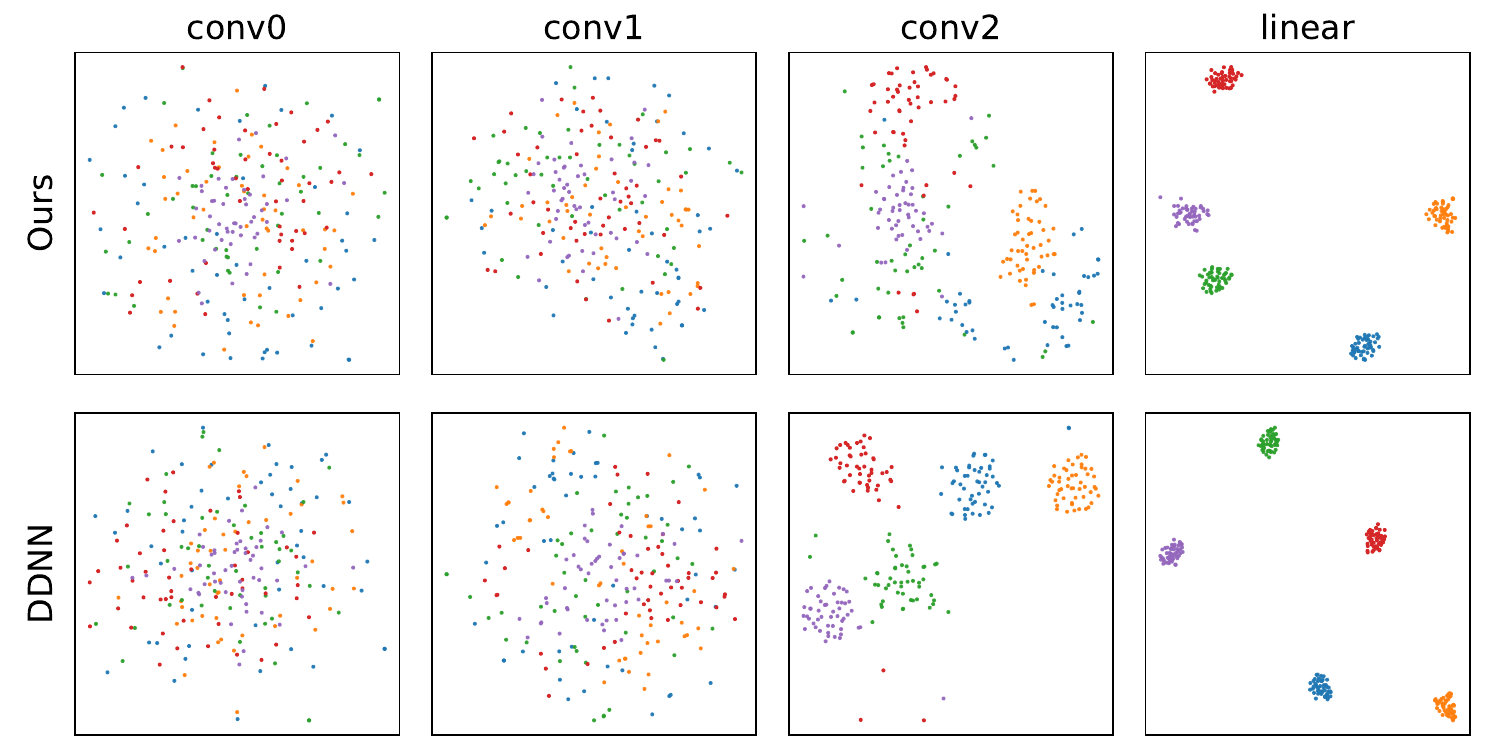}}}
    \vspace{-0.3cm}

    \subfigure[No Calibration (Original Cross-Entropy)]{
    {\includegraphics[width=0.48\textwidth,height=0.22\textwidth]{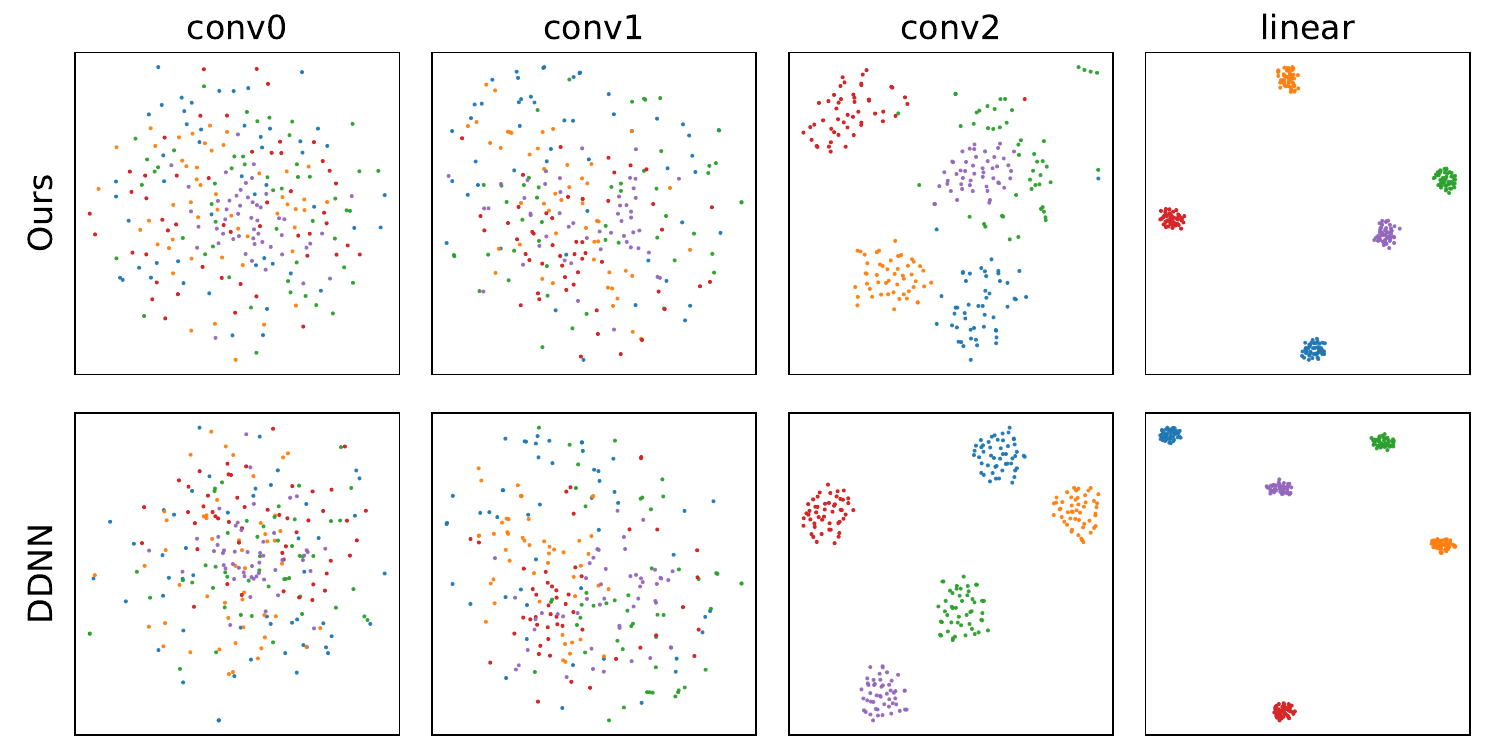}}}
    
    \caption{T-SNE projections of feature vectors from each layer of a 4-block ConvNet trained with mixup, label smoothing, focal loss, and the original cross-entropy on distilled CIFAR10. In each training method, applying our proposed MDT (Ours) helps the network encode more source information in intermediate layers, as visualized by the rich features not separated until the last layer. The original DDNN poorly encodes source information, as shown by the feature projections already separated in layer conv2.}
    \label{fig:supp_proj}
    \vspace{-0.3cm}
\end{figure}

\noindent \textbf{Comparison of Effects of Different During-Training Calibration Methods on DDNNs' Encoding Capacity.}
We provide more visualizations of projections of intermediate feature vectors obtained from DDNNs trained with different during-training calibration methods. The methods we use are mixup, focal loss, and label smoothing, in addition to the original training with cross-entropy loss. We can see in Figure~\ref{fig:supp_proj} that our proposed during-training calibration MDT alleviates the issue of concentrate features for all the traditional methods used, giving better encoding potentials of DDNNs for transfer learning tasks, which leads to more calibratable DDNNs.

\begin{figure}[!t]
\vspace{-0.2cm}
\setlength{\abovecaptionskip}{-0.0cm}
\setlength{\belowcaptionskip}{-0.0cm}
    \caption{Ours (MTS) better calibrates DDNNs across different IPC in MTT. Left: CIFAR10. Right: CIFAR100.}
    \subfigure{
    {\includegraphics[width=0.23\textwidth,height=0.15\textwidth]{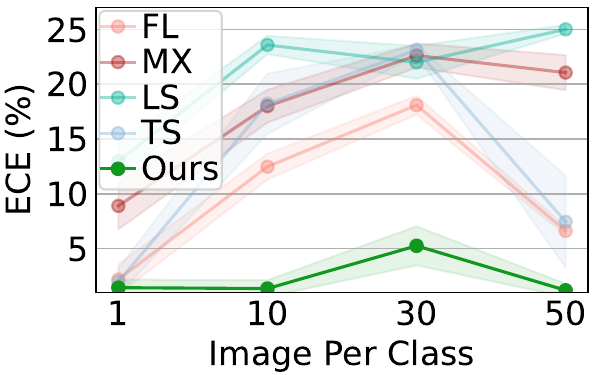}}} \vspace{-0.1cm}
    \subfigure{
    {\includegraphics[width=0.23\textwidth,height=0.15\textwidth]{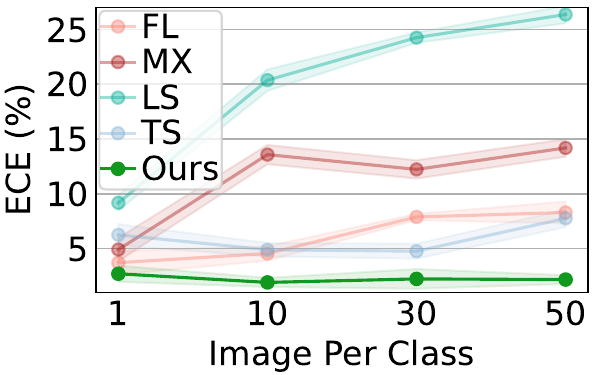}}}
    \label{fig:supp_ipc_ece}
    \vspace{-0.5cm}
\end{figure}

\noindent \textbf{More Results on CIFAR100: ECE on different IPCs, max logits.}
We show in Figure~\ref{fig:supp_ipc_ece} that our MTS outperforms others in ECE on different IPCs. 
In the main paper, we mainly present IPC = 10 on Tiny-ImageNet \& Subsets with MTT, 10 on CIFAR100 with DC/DSA (released), and 50 on others. These DD settings have higher accuracy and would better represent real-world settings.

\begin{figure}[!t]
\vspace{-0.1cm}
    \centering
    \caption{The more calibratable FDNN outputs more evenly distributed logits, while the less calibratable DDNN outputs a more concentrated logit distribution.}
    \subfigure[CIFAR 100.]{
    {\includegraphics[width=0.23\textwidth,height=0.21\textwidth]{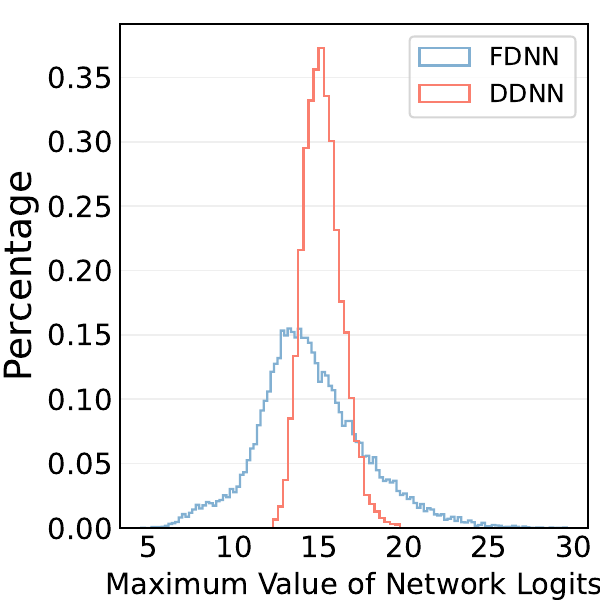}}}
    \label{fig:supp_max_logit_100}
    \vspace{-0.4cm}
\end{figure}

We also provide visualization of maximum logits of DDNN on original MTT in Figure~\ref{fig:supp_max_logit_100}, in addition to the results on CIFAR10 in our main paper.

\section{Conclusion}

In this paper, we find for the first time that networks trained on distillation data are not calibratable and have poor encoding ability because the distillation process focuses on the classification task while discarding other semantically meaningful information. Our proposed methods, namely Masked Distillation Training during training and Masked Temperature Scaling after training, effectively alleviate these limitations and make the DDNNs recalibrated. 

In future work, we will look for better distillation methods that retain most of the source information and lead directly to calibratable networks. In addition, beyond calibrating DDNNs on in-distribution data, we will rethink DDNNs in terms of more general reliability, i.e., out-of-distribution detection, robust generalization, and adaptation, which are important properties for the safety of DDNN applications.

{\small
\bibliographystyle{ieee_fullname}
\bibliography{arxiv}
}

\clearpage
\appendix

\section{Distillation Backbones}

\subsection{Datasets and Networks}
Following ~\cite{zhao2020dataset, zhao2021dataset, cazenavette2022dataset, deng2022remember}, we use a ConvNet with 3 blocks for CIFAR10 and CIFAR100~\cite{krizhevsky2009learning}, a ConvNet with 4 blocks for Tiny-ImageNet~\cite{le2015tiny}, and a ConvNet with 5 blocks for Nette (a subset of ImageNet)~\cite{Howard_Imagenette_2019}. Each block in the ConvNets contains a 3 $\times$ 3 convolutional layer with 128 channels, followed by instance normalization~\cite{ulyanov2016instance}, ReLU~\cite{nair2010rectified} and a 2 $\times$ 2 average pooling layer with stride 2. We apply Kornia ZCA~\cite{riba2020kornia} on CIFAR10 and CIFAR100 for distillation backbones~\cite{zhao2020dataset, zhao2021dataset, cazenavette2022dataset}. We pick the ConvNet in each distillation backbone because it gives the best distillation performance while keeping the distillation process under an acceptable time and computational budget.

\section{Additional Experiments}

\subsection{Details in Masked Temperature Scaling}
We sample from all the distilled data we have as the validation set to update the temperature parameter $T$ in our proposed Masked Temperature Scaling. Instead of sampling from all the shuffled data at once, we perform a per-class sampling such that there is no missing class or over-sampled class, which is especially important for distillation settings that aim for aggressive compression rates such as image-per-class $\leq$ 10. The traditional temperature scaling~\cite{guo2017calibration} separates all the data available into a training set and a validation set and uses the validation set only for updating $T$. This separated use of the distilled data is not applicable when image-per-class = 1. Moreover, a data split of 10\% can hurt training accuracy by as much as 1.68\% on the Nette subset of ImageNet, while our proposed during-training calibration method (MDT) only hurts accuracy by 0.24\%, as reported in Table~\ref{tab:ts_acc_table}. In addition, our proposed after-training method Masked Temperature Scaling keeps original training accuracy and achieves better calibration results than temperature scaling as reported in our main text.

\begin{table}[!t]
    \addtolength{\tabcolsep}{0pt}
    \renewcommand{\arraystretch}{0.9}
    \centering
    \caption{Accuracy (\%) drops by as much as 1.68\% when training with 90\% of distilled Nette (a subset of ImageNet). The rest 10\% is used in temperature scaling (TS). Our proposed after-training MTS (\colorbox[gray]{0.8}{shadow}) keeps the original accuracy. Our proposed during-training MDT (\colorbox[gray]{0.9}{shadow}) keeps a higher accuracy than that of dropping 10\% of training data for TS. We use MTT~\cite{cazenavette2022dataset} as the distillation backbone.}
    \label{tab:ts_acc_table}

    \scalebox{0.83}{

    \setlength{\aboverulesep}{0pt}
    \setlength{\belowrulesep}{0pt}
    \setlength{\extrarowheight}{0.8ex}

    \begin{tabular}{l|>{\columncolor[gray]{0.8}}c|c>{\columncolor[gray]{0.9}}c}
    \toprule
    \multicolumn{1}{c|}{Dataset}         & Full, MTS (Ours)        & TS (10\%)                 & MDT (Ours)                      \\ \hline
    \multicolumn{1}{l|}{CIFAR10}         & 70.48 $\pm$ 0.2          & 69.78 $\pm$ 0.5         & 69.98 $\pm$ 0.4        \\
    \multicolumn{1}{l|}{CIFAR100}        & 47.47 $\pm$ 0.2          & 47.10 $\pm$ 0.2      & 46.21 $\pm$ 0.4                 \\
    \multicolumn{1}{l|}{Tiny ImageNet}   & 27.76 $\pm$ 0.2          & 27.35 $\pm$ 0.2         & 27.62 $\pm$ 0.4        \\
    \multicolumn{1}{l|}{ImageNette}      & 63.04 $\pm$ 1.3          & 61.36 $\pm$ 1.6         & 62.80 $\pm$ 1.2                 \\ \bottomrule
    \end{tabular}
    }
    \vspace{-0.5cm}
\end{table}

\begin{table}[!t]
    \addtolength{\tabcolsep}{-2pt}
    \renewcommand{\arraystretch}{1}
    \centering
    \caption{ECE (\%) of different calibration methods on FDNNs. With a low masking ratio $r$, our results (\colorbox[gray]{0.9}{shadow}) are comparable to temperature scaling and most of the time beats other methods. As our method is specifically designed for DDNNs, in the case of FDNNs where traditional methods are suitable, we can simply convert our method to temperature scaling by setting $r$ to 0.}
    \label{tab:fd_ece_table}
    \scalebox{0.9}{

    \setlength{\aboverulesep}{0pt}
    \setlength{\belowrulesep}{0pt}
    \setlength{\extrarowheight}{.75ex}
    
    \begin{tabular}{l|c|cccc>{\columncolor[gray]{0.9}}cc}
    \toprule
    \multicolumn{1}{c|}{Dataset} & Raw & TS   & MX           & LS & FL      & MTS               \\ 
    \hline
    CIFAR10                           & 4.50       & 0.99 & 14.80 & 11.85 & 1.78 & 2.67                                \\
    CIFAR100                          & 13.05       & 1.41 & 10.69          &   7.17     &  3.49        & 1.84                                  \\
    Tiny ImageNet                      & 22.26       & 4.95          & 6.34          &   3.29      & 12.55         & 4.93                         \\
    ImageNette                    & 10.90       & 2.81 & 11.22          & 22.24 &   5.21       & 2.87                             \\ \bottomrule
    \end{tabular}}
    \vspace{-0.4cm}
\end{table}

\begin{figure}[!b]
    \vspace{-0.2cm}
    \centering
    {\includegraphics[width=0.5\textwidth,height=0.3\textwidth]{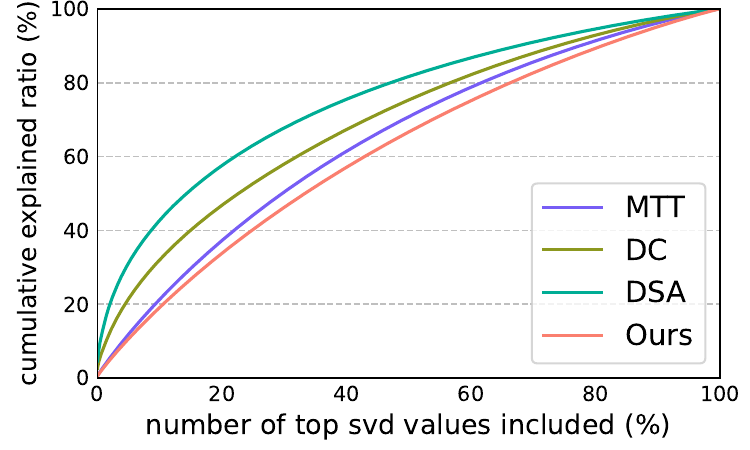}}
    \caption{Cumulative \textit{explained ratio}, i.e., percentage of top singular values to $\sum{\operatorname{diag}\left(\Sigma\right)}$ in SVD decomposition of distilled CIFAR10 from different distillation backbones. Ours (red) grows at the most steady rate, indicating its evenly distributed information, compared to others with condensed information.}
    \label{fig:supp_svd}
    \vspace{-0.2cm}
\end{figure}

\subsection{More Results on SVD of Distilled Data and Full Data}

As we discussed in our main text, distilled data contain more concentrated information that easily gets grouped by algorithms such as SVD. We here illustrate the cumulative explained ratio of top singular values of data distilled by different backbones. We expect that concentrated information leads to a curve skewed to the top left and evenly distributed information leads to a smooth curve close to the diagonal. This will show how much each component corresponding to the singular values in $\Sigma$ contributes to the data reconstruction. As shown in Figure \ref{fig:supp_svd}, the cumulative explained ratio given by ours grows at the most steady rate, showing that our method produces more evenly distributed information in distilled data compared to the overly condensed information in other distillation backbones. As we concluded in our main text, this serves as a regularization to the distillation process such that it cannot discard too much information that is unrelated to the classification task but semantically meaningful for other tasks, leading to more calibratable networks trained on the resulting distilled data.

\subsection{Performance Analysis of FDNNs}

We further test MTS on the more calibratable FDNNs. We calibrate networks trained on the full CIFAR10, CIFAR100, TinyImageNet, and Nette subset of ImageNet. We report the mean of 2 runs due to limited computational resources. As reported in Table~\ref{tab:fd_ece_table}, our method performs comparably with existing well-developed methods. In realistic settings with a large amount of training data, we can set the masking ratio $r$ to 0, which converts the MTS back to normal temperature scaling.

\end{document}